\documentclass[letterpaper, 10 pt, conference]{ieeeconf}
\IEEEoverridecommandlockouts
\usepackage{multicol}
\usepackage{epsfig}
\usepackage{color}
\usepackage{graphicx}
\usepackage{multirow}
\usepackage[table]{xcolor}
\usepackage{url}
\usepackage[rflt]{floatflt}
\usepackage{subfigure}
\usepackage{textcomp}
\usepackage[percent]{overpic}
\usepackage{wrapfig}
\usepackage{xspace}
\usepackage{booktabs}
\usepackage{tabularx}
\usepackage{adjustbox}
\usepackage{amsmath}
\usepackage[colorinlistoftodos,prependcaption,textsize=tiny,disable]{todonotes}
\usepackage{mathtools}
\usepackage{bm}
\usepackage{cuted}    
\usepackage{caption}  
\usepackage{array}
\usepackage{lipsum}
\usepackage{hyperref}
\usepackage{amsfonts}

\title{\LARGE \bf
\textsf{GRIP}: A \textsf{G}eneral \textsf{R}obotic \textsf{I}ncremental \textsf{P}otential Contact Simulation Dataset for Unified Deformable-Rigid Coupled Grasping}

\author{Siyu Ma$^{*1}$, Wenxin Du$^{*1}$, Chang Yu$^{*1}$, Ying Jiang$^{*1}$, Zeshun Zong$^{1}$, Tianyi Xie$^{1}$, Yunuo Chen$^{1}$, \\ Yin Yang$^{3}$, Xuchen Han$^{2}$, Chenfanfu Jiang$^{1}$
\thanks{* equal contribution.}
\thanks{$^{1}${\tt\footnotesize siiyuma@outlook.com,\{wenxindu,changyu1,yingjiang,
zeshunzong,tianyixie77,yunuoch,cffjiang\}@ucla.edu}, AIVC Laboratory, UCLA, USA.}
\thanks{$^{2}${\tt\footnotesize xuchen.han@tri.global}, Toyota Research Institute, USA.}
\thanks{$^{3}${\tt\footnotesize yin.yang@utah.edu}, University of Utah, USA.}
}

\newcommand{\vf}[1]{{\bm{#1}}}
\newcommand{\mf}[1]{{\mathbf{#1}}}

\begin{document}
\maketitle

\thispagestyle{empty}
\pagestyle{empty}
\begin{strip}
\begin{minipage}{\textwidth}\centering
\vspace{-22mm}
\includegraphics[width=\textwidth]{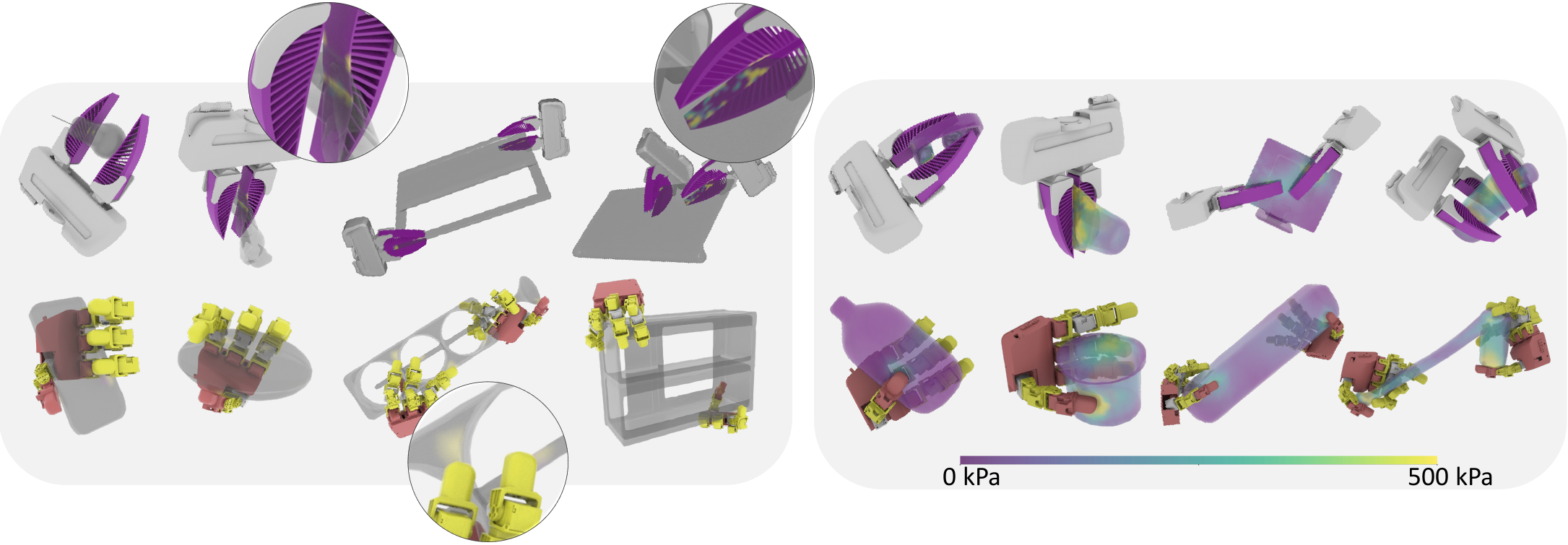}
\captionof{figure}{\textbf{GRIP} is a large-scale and universal grasping dataset that comprises 100K high-quality grasps across diverse scenarios. It includes soft UMI grippers (\textbf{top}) and rigid LEAP Hands (\textbf{bottom}), interacting with rigid (\textbf{left}) and soft (\textbf{right}) objects under both unimanual and bimanual settings. Our dataset captures a wide variety of object shapes, sizes, and materials, along with rich object and gripper deformations and stress distribution by an IPC simulator with high-fidelity frictional contact.}
\vspace{-2mm}
\label{fig:teaser}
\end{minipage}
\end{strip}

\begin{abstract}
Grasping is fundamental to robotic manipulation, and recent advances in large-scale grasping datasets have provided essential training data and evaluation benchmarks, accelerating the development of learning-based methods for robust object grasping. However, most existing datasets exclude deformable bodies due to the lack of scalable, robust simulation pipelines, limiting the development of generalizable models for compliant grippers and soft manipulands. To address these challenges, we present GRIP, a General Robotic Incremental Potential contact simulation dataset for universal grasping. GRIP leverages an optimized Incremental Potential Contact (IPC)-based simulator for multi-environment data generation, achieving up to 48× speedup while ensuring efficient, intersection- and inversion-free simulations for compliant grippers and deformable objects. Our fully automated pipeline generates and evaluates diverse grasp interactions across 1,200 objects and 100,000 grasp poses, incorporating both soft and rigid grippers. The GRIP dataset enables applications such as neural grasp generation and stress field prediction. We release GRIP to advance research in robotic manipulation, soft-gripper control, and physics-driven simulation at: \href{https://bell0o.github.io/GRIP/}{https://bell0o.github.io/GRIP/}. 
\end{abstract}

\section{INTRODUCTION}
Grasping is pivotal to robotics, serving as the cornerstone of nearly all manipulation tasks.
However, acquiring reliable grasping data for robotics training faces challenges such as object variability, occlusions, and generalization to unseen objects. The emergence of large-scale grasp datasets, such as GraspNet-1Billion~\cite{graspnet_1billion} and DexGraspNet~\cite{wang2023dexgraspnet}, has significantly advanced research in this area, providing rich data to improve grasp prediction, facilitating applications across industrial automation, household robotics, and healthcare.

More recently, developments in soft grippers, such as the UMI gripper~\cite{umi_gripper} and the Bubble gripper~\cite{bubble_gripper}, have further advanced grip stability. Due to their compliance, soft grippers are particularly suitable for grasping fragile objects such as fruits, food, and biological tissues, where preventing damage is crucial.

Despite recent progress in soft grippers, grasp datasets primarily emphasize rigid grippers and manipulands, with none specifically dedicated to soft grippers. The DefGraspSim~\cite{huang2021defgraspsim} dataset remains the only notable attempt to include deformable manipulands, but its limited scale (34 objects and 6,800 grasp evaluations) falls short of supporting the development of generalizable grasp prediction models. The scalability of real-world datasets for soft grippers and deformable manipulands is limited by their high degrees of freedom (DoFs). This complexity makes it difficult to track deformations accurately, especially when occlusions occur during grasping. Consequently, simulations are often favored over real-world data.

An ideal physical simulator for soft grippers and deformable objects must satisfy three key criteria. (1) \emph{Robustness}: It should accurately handle soft-soft, soft-rigid, and rigid-rigid interactions across diverse objects and dynamic conditions without introducing critical artifacts. (2) \emph{Efficiency}: It should be parallelized to enable large-scale dataset generation. (3) \emph{Accuracy}: It must provide precise physical dynamics and realistic frictional contact. However, existing simulators struggle with reliable contact resolution in complex geometries, such as strings, cones, or thin surfaces, often producing tunneling artifacts that distort physical interactions and deviate from real-world behavior. Moreover, most FEM-based soft-body simulators are prone to element inversion under large deformations, further compromising simulation accuracy. 

The recently proposed Incremental Potential Contact (IPC) method~\cite{li2020incremental} serves as a remedy, effectively addressing large-deformation and frictional contact problems for soft bodies within a variational framework, while guaranteeing intersection- and inversion-free solutions. Recent extensions and developments, such as \cite{chen2022midas, zema}, have integrated IPC and Affine Body Dynamics (ABD)~\cite{abd} into a unified framework with multibody and articulated dynamics~\cite{chen2022unified}. However, these approaches either lack user-friendly interfaces for robotic applications, or are not optimized for large-scale, multi-environment simulations, resulting in inefficiencies in robotic applications.

To address the aforementioned challenges, we present \textbf{GRIP}: a \textbf{G}eneral \textbf{R}obotic \textbf{I}ncremental \textbf{P}otential contact simulation dataset for universal grasping. Our key contributions include the following: 
\begin{enumerate} 
\item \textbf{A high-performance robust IPC-based simulator} optimized for large-scale, multi-environment dataset generation, incorporating the four core features outlined earlier. Our simulator achieves a 48× speedup when running 400 parallel environments compared to single-environment sequential simulations.
\item \textbf{A fully automated pipeline for grasp generation, simulation, and evaluation.} The pipeline generates synthesized grasps and simulation results such as gripper and manipuland deformation and stress distribution. It also supports various types of grippers (parallel and dexterous, rigid and soft) and manipulands (rigid and soft) for both uni- and bi-manual grasp settings. 
\item \textbf{The GRIP dataset}, a large-scale and diverse grasping dataset containing \textbf{1200 objects and 100K grasp poses}, featuring both soft UMI grippers~\cite{umi_gripper} and rigid LEAP Hand grippers~\cite{shaw2023leap}.
\item \textbf{Practical applications of the GRIP dataset}, showcasing its effectiveness in neural grasp generation and stress field prediction for soft manipulands.
\end{enumerate}

\begin{figure*}[h]
\centering
\includegraphics[width=\linewidth]{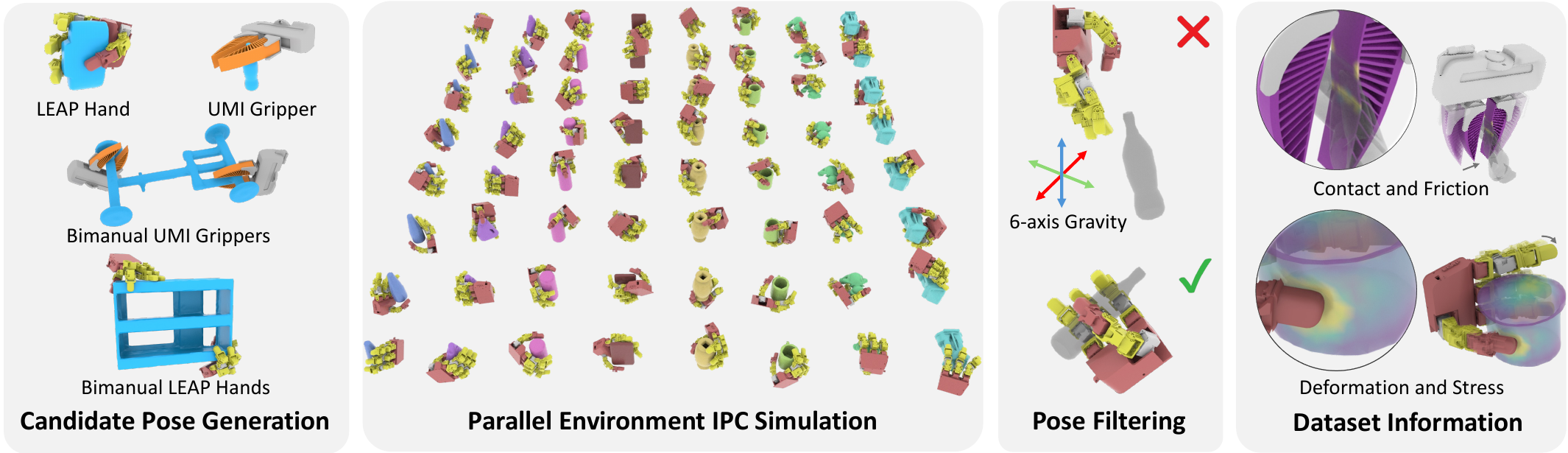}
\caption{\textbf{Dataset Generation Pipeline}: Candidate grasps are first synthesized using GPD~\cite{gpd} and DexGraspNet~\cite{wang2023dexgraspnet}. These candidate poses are then evaluated in a parallelized IPC simulator, where 6-axis gravity is applied to assess grasp stability. The resulting dataset captures contact and friction data, along with detailed deformation and stress information for soft manipulands and grippers. It provides a comprehensive record of contact interactions and material responses across all simulation time steps throughout the entire grasping trajectory.}

\vspace{-2mm}
\label{fig:pipeline}
\end{figure*}

\section{RELATED WORK}
\subsection{Dataset for Hand Grasping}
Existing grasp datasets for robotic grippers primarily focus on rigid gripper–rigid object interactions. Many center on parallel grippers \cite{eppner2021acronym, morrison2020egad}, object-centric relationships \cite{zhang2022regrad}, or multimodal sensing~\cite{wang2019multimodal}. These grippers are typically composed of rigid links connected via joints. For example, \cite{graspnet_1billion, grasp_anything} provide grasp poses for rigid parallel grippers. Dexterous rigid grippers, which offer greater agility and versatility due to their increased degrees of freedom (DoFs), enable a wider range of human-like grasping motions. \cite{wang2023dexgraspnet, zhang2024graspxl} synthesize dexterous poses for ShadowHand and Allegro, whereas \cite{li2023gendexgrasp} generates grasp poses for five different dexterous hands. \cite{casas2024multigrippergrasp} introduced a large-scale synthetic dataset covering 11 common types of rigid grippers. In addition to single grippers, \cite{shao2024bimanual} introduces a large-scale synthetic dataset focusing on bimanual grasping with heavy or large objects.

Recently, efforts have also been made to collect grasp poses for soft objects. For instance, \cite{huang2021defgraspsim} offers simulated deformation and stress distribution for 34 deformable manipulands grasped by rigid grippers using the Isaac Gym \cite{isaac_gym} simulator. Soft grippers, such as the UMI gripper \cite{umi_gripper} and the bubble gripper \cite{bubble_gripper}, can offer more stable grasps due to their superior compliance. However, none of the existing grasp datasets include soft grippers, which limits the progress of their downstream tasks.

\subsection{Grasp Generation}
Most prior works formulate grasp synthesis as an optimization problem. For instance, GraspIt!~\cite{miller2004graspit} generates grasp poses using annealing-based search algorithms. However, its simplified optimization objectives and search strategies limit both the efficiency and diversity of generated grasp poses~\cite{diff_fc, wang2023dexgraspnet}. To address this, \cite{diff_fc} introduces a differentiable force-closure term, enabling force-closure-aware, optimization-based grasp synthesis, thus yielding more diverse grasp poses. Building on this approach, DexGraspNet~\cite{wang2023dexgraspnet} enhances gripper-object interaction by incorporating a robust, differentiable object-to-hand penetration energy term. Additionally, DexGraspNet significantly improves grasp success rates by leveraging the Isaac Gym simulator~\cite{isaac_gym} to validate and filter grasp poses. This method has facilitated the creation of a large-scale dexterous grasp dataset for ShadowHand. Xu et al.~\cite{xu2024dipgrasp} proposed a gradient descent-based optimizer with a surface-matching metric. In addition to meshes, it supports point clouds as object geometry representation, enhancing real-world applicability, since point clouds are more accessible in practical scenarios. Nevertheless, their method requires intricate parameter tuning to generate diverse grasp poses.

Recently, data-driven methods have advanced grasp generation through diffusion-based dexterous grasping \cite{lu2024ugg}, reinforcement learning \cite{christen2022d}, and grasp proposal networks for both 6-DOF \cite{pcd_neural_graspgen_1, pcd_neural_graspgen_2} and dexterous grasping \cite{pcd_neural_graspgen_3}. These approaches predict grasp poses based on object geometry and efficiently generate diverse grasps in parallel, making them well-suited for real-time applications. Moreover, many studies~\cite{lu2024ugg, pcd_neural_graspgen_1, pcd_neural_graspgen_2, pcd_neural_graspgen_3} support point cloud representations of grasped objects, further improving their practicality for real-world deployment.

\subsection{Robot Simulation}
Mainstream robotic deformable body simulation algorithms often adopt the Material Point Method (MPM) for tasks such as tactile sensing \cite{si2024difftactile}, soft robot control \cite{hu2019chainqueen}, and soft object manipulation \cite{zeshun_mpm_rigid}, or use the Finite Element Method (FEM)~\cite{fem_robot_1, huang2022defgraspsim} with various dimensional topologies~\cite{zema, yu2023diffclothai}. MPM models an object as a set of particles without explicit topology, effective for handling large topology changes. However, it relies on a background Eulerian grid, which introduces numerical diffusion artifacts and incurs high computational costs as grid resolution increases \cite{mpm_numerical_diffusion, wang2020massively}. FEM represents objects as structured meshes, such as triangular elements for surface meshes and tetrahedral elements for volume meshes, and yields great accuracy when objects undergo small deformations. Nevertheless, for most FEM-based methods, large deformations will lead to element inversion, resulting in simulation failures \cite{espinosa1998adaptive, irving2004invertible}. Further, unlike MPM, FEM requires explicit collision handling, which is commonly formulated as a Linear Complementarity Problem (LCP) \cite{lcp_contact_1, lcp_contact_2} or a convex optimization problem \cite{todorov2012mujoco, convex_contact_1}. LCP-based methods are hindered by their NP-hardness and ill-conditioning issues, while convex-optimization-based approaches face challenges in extreme cases, such as fast-moving non-convex objects and extremely thin objects.

Recently, Li et al. proposed Incremental Potential Contact (IPC)~\cite{li2020incremental}, a FEM-based method that accurately and robustly handles extreme contact scenarios. IPC formulates frictional contact within FEM as an optimization problem, incorporating barrier augmentation, continuous collision detection (CCD), and a Projected Newton optimization scheme with a step-size filter to enforce intersection- and inversion-free constraints \cite{li2024physics}. Despite its high accuracy in robotic simulations, including dataset generation \cite{ipc_graspsim}, tactile modeling \cite{du2024tacipc}, and general-purpose simulation \cite{fernandez2024stark}, the computational inefficiency of IPC limits its applicability to large-scale simulations, hindering its adoption in large data generation and parallel reinforcement learning training. 

Our work extends IPC to a large-scale parallel environment simulation, improving its efficiency and scalability through parallel computing. We construct a large-scale grasp pose dataset using optimization-based pose synthesis combined with a multi-environment parallel IPC simulator.

\section{GRIP DATASET GENERATION}

We outline the complete methodology for our dataset generation. Our approach utilizes a GPU-parallelized, multi-environment high-performance IPC simulator (Section \ref{sec:ipc_simulator}) to enable large-scale dataset generation with strict non-penetration guarantees, providing precise contact and stress information. The overall data generation pipeline is illustrated in Fig. \ref{fig:pipeline}. We first convert grippers and objects into simulation-ready assets (Section \ref{sec:data_prep}). Next, we use grasp synthesis algorithms to generate candidate grasp poses (Section \ref{sec:grasp_generation}), which are evaluated using the IPC simulator. Afterwards, unstable grasps are filtered out (Section \ref{sec:grasp_validation}).

\subsection{IPC Simulator with Parallel Environment}
\label{sec:ipc_simulator}
A straightforward approach to implementing a fully GPU-parallelized IPC simulator for multi-environment simulation is to stack all environments into a single system within a shared scene. This method is inefficient and prone to failures for several reasons: \emph{(i)} originally independent simulations may interfere, causing unintended collisions; \emph{(ii)} to ensure non-penetration, the global Newton step size is constrained by the smallest non-colliding step size across all environments, reducing overall efficiency; \emph{(iii)} varying convergence iterations across different environments lead to wasted computation time for already converged environments; \emph{(iv)} a failure in one environment causes the entire system to fail.

To address these challenges, we optimize the solver pipeline to support large-scale parallel simulation. First, we introduce environment isolation during collision detection and step size filtering, allowing each environment to evolve independently with its own optimization step size. Second, we track individual stopping criteria (e.g., relative error in Newton iterations) for each environment. Once an environment meets its stopping criteria, it is frozen and excluded from further computations, freeing up resources for the remaining active environments. Third, we introduce a failure detection stage in the simulation pipeline to identify and exclude failed environments from further computation, isolating them from the overall system to ensure robustness. In addition, we incorporate parallel computing in linear solves to boost simulation efficiency. The performance of our parallel simulator is evaluated in Section \ref{sec:exp_parallel_env}.

\subsection{Grippers and Objects}
\label{sec:data_prep}
\paragraph{Gripper Modeling} 
Our dataset covers both UMI grippers~\cite{umi_gripper} and LEAP Hands~\cite{shaw2023leap}. For simulation, we tetrahedralize the soft finger meshes of UMI grippers using TetWild \cite{hu2020fast}, setting a tight envelope tolerance of $\epsilon_e=5\times10^{-4}$ to preserve fine surface geometry details. 
The Young's modulus ($E_{\text{UMI}}=9.4\times10^6$ Pa) and dry friction coefficient ($\mu_{\text{UMI}}=3.5$) of the UMI gripper fingers are set to match their real-world counterparts \cite{umi_gripper}.
For LEAP Hands, leveraging IPC's robustness for handling complex geometries,
we use their high-resolution visual meshes directly as collision meshes for precise contact resolution in simulation. To generate the contact point candidates for DexGraspNet synthesis, we heuristically sample points evenly across the contact regions of the hand's palms and fingers.

\paragraph{Object Preparation} 
We select 800 objects from the DexGraspNet dataset~\cite{wang2023dexgraspnet} for single-hand grasping via stratified sampling based on label categories and their object counts. The subset matches the original distribution, with at least one object included from underrepresented labels to ensure diversity. For bimanual grasping, we retain 400 large objects from the single-hand set and add additional 400 objects from the PartNet-Mobility dataset~\cite{xiang2020sapien}, totaling 800 large-scale objects. To prepare for the simulation, we preprocess the surface meshes to be watertight and tetrahedralize them for soft body simulations using TetWild. We also apply domain randomization, sampling Young's modulus, friction coefficient, and object scaling to enhance data diversity.

\subsection{Grasp Pose Synthesis}
\label{sec:grasp_generation}
We present our grasp pose synthesis algorithms for the UMI gripper and the LEAP Hand. The synthesized grasp poses serve as candidate configurations, which are subsequently validated and filtered with our simulator.

For the UMI gripper, we use the GPD method~\cite{gpd}, a simple yet effective solution for parallel grippers. Since IPC simulation requires an intersection-free initial configuration, we discard a small fraction of the synthesized grasp poses where the gripper penetrates the object.

For the LEAP Hand, we adopt DexGraspNet~\cite{wang2023dexgraspnet}, replacing the ShadowHand with the LEAP Hand, and incorporate a normal alignment energy term $E_{\text{normal}}$ defined as 
\begin{equation}
E_{\text{normal}} = \sum_{i=1}^{n_{c}} ((\mf{R}_i^{\text{h}}\vf{n}_i^{\text{h}})^t\vf{n}^{\text{o}}_i+1)^2,   
\end{equation}
from \cite{xu2024dipgrasp} into the optimization process. This modification encourages alignment between the contact normals of the LEAP Hand and the object, and significantly improves grasp success rates. Here $n_c$ is the number of contact points, $\vf{n}_i^{\text{h}}$ represents the normal vector of the $i$-th hand contact point in its associated link's local frame $\mf{T}_i$, $\mf{R}_i$ denotes the rotation from $\mf{T}_i$ to the world frame, and $\vf{n}_i^{\text{o}}$ is the normal vector of the $i$-th object contact point in the world frame. 

Although DexGraspNet efficiently generates grasp poses, minor penetrations between the gripper and object often occur, violating IPC's prerequisite for intersection-free initial configurations. To address this, we apply inverse kinematics to adjust fingertip positions along the object's surface normals, ensuring contact distances exceed IPC's contact threshold $\hat{d}$. For GRIP dataset generation, we set $\hat{d} = 10^{-3}$~m. This method effectively resolves most initial penetrations, with any remaining penetrated poses discarded.

For bimanual grasp poses, we synthesize $n_{\text{target}}$ bimanual grasp candidate poses for each object by composing unimanual-grasp poses. First, we apply k-means clustering to group left and right gripper candidate poses into $k = 26$ clusters, using the center positions of contact points as grouping keys. These clusters form $k^2$ potential pairs of left-hand and right-hand pose groups. Next, we select the top $r_1 k^2$ cluster pairs where the Euclidean distance between the group-averaged contact position centers of the left and right hands is the largest. For GRIP dataset generation, we set $r_1=0.25$. Within each selected cluster pair, we retain all possible left-hand and right-hand pose combinations, denoting the total number of such bimanual grasp poses as $n_{\text{filtered}}$. Finally, we compute an approximated force closure metric for each bimanual grasp pose \cite{diff_fc, wang2023dexgraspnet} and retain the bottom $r_2=\frac{n_{\text{target}}}{n_{\text{filtered}}}$ fraction within each cluster pair, yielding approximately $n_{\text{target}}$ final bimanual grasp candidates.

\begin{figure}[t]
\centering
\includegraphics[width=1.0\linewidth]{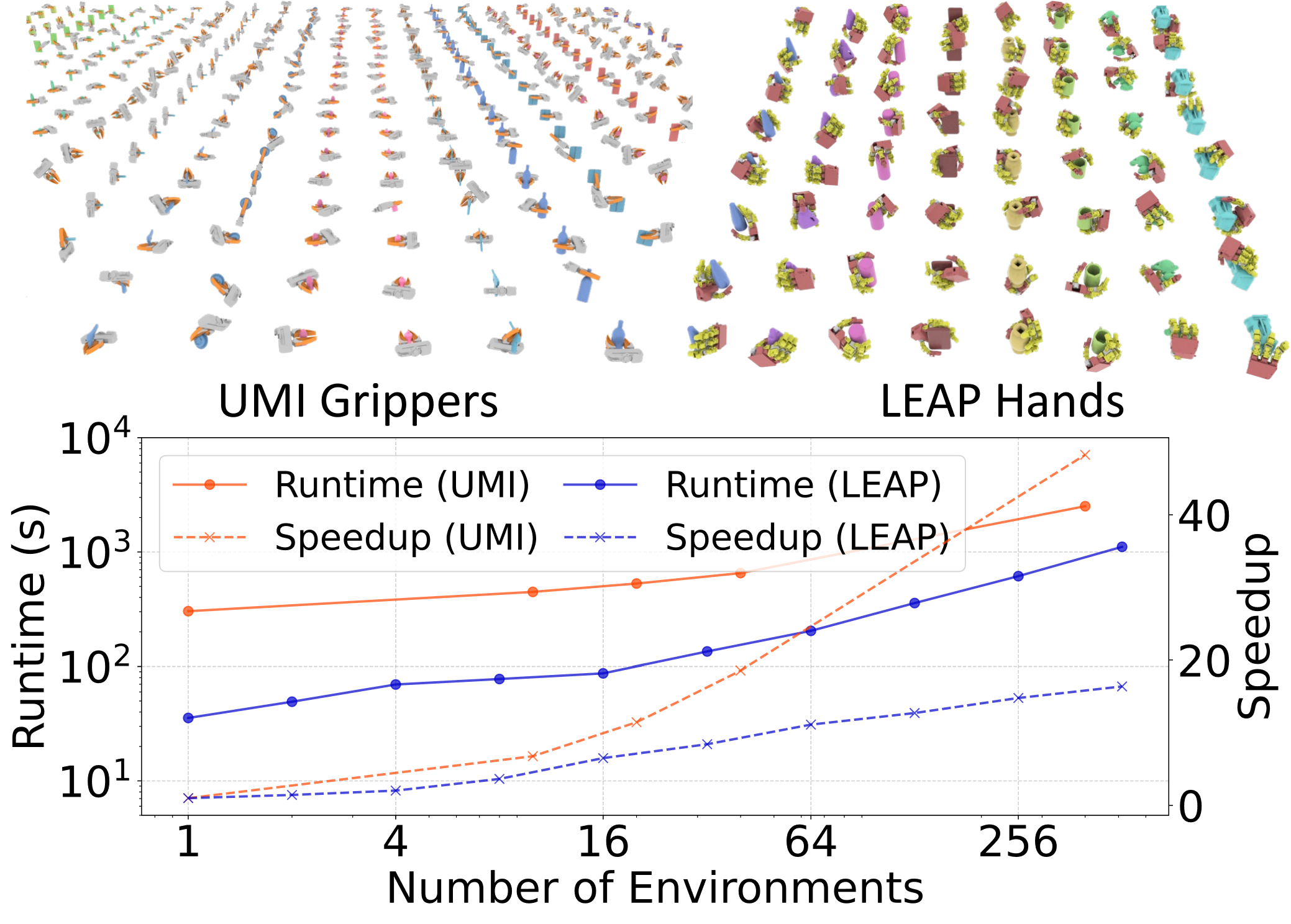}
\caption{\textbf{Performance Analysis of Our Parallel IPC Simulator}: Runtime profiling for unimanual UMI grippers and LEAP Hands under varying numbers of parallel environments.
We compare our runtime and speedup with sequential execution, demonstrating the strong scalability.}
\vspace{-3mm}
\label{fig:scalability}
\end{figure}

\subsection{Grasp Validation}
\label{sec:grasp_validation}
We use our parallel IPC simulator to filter out failed grasps and record simulation data for successful ones. The time step size is set to $\Delta t = 0.01$ seconds. The contact stiffness is $\kappa = 3 \times 10^6~\text{kg}\cdot \text{s}^{-2}$, the contact distance threshold is $\hat{d} = 10^{-3}$ m, and the velocity magnitude threshold for dynamic-static friction transition is $\epsilon_{v} = 10^{-3}$ m/s, ensuring precise frictional contact handling. Simulations are run with FP64 precision. The Newton optimization terminates when the relative tolerance $\epsilon_r = 10^{-3}$ is reached. If optimization fails to converge within $N_{\text{iters}} = 100$ iterations, the environment is marked as failed.

\label{sec:exp1_real2sim}
\begin{figure}[t]
    \centering
    \includegraphics[width=1\linewidth]{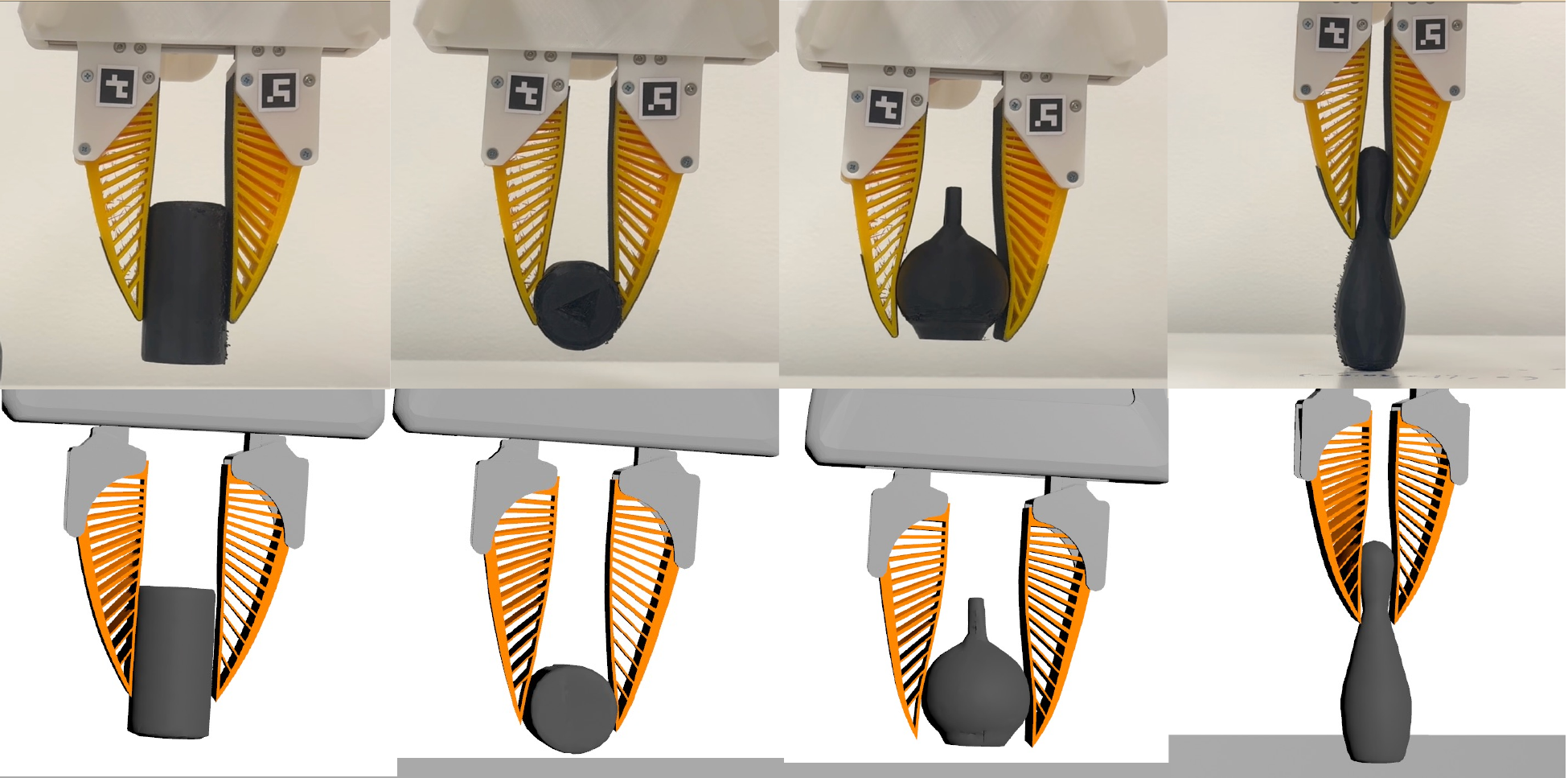}
    \caption{\textbf{Comparison of Real-World and Simulated Grasps}: Grasps of four 3D-printed manipulands in the real world (\textbf{top}) and in simulations (\textbf{bottom}), showing strong consistency between real-world and simulated results, demonstrating the high fidelity of our simulator.}
    \vspace{-3mm}
    \label{fig:real2sim_exp1}
\end{figure}

At the start of each trajectory, gravity is disabled to prevent the object from falling before being grasped. To simulate contact-aware grasping, a finger's motion is halted once its contact force exceeds 50 N. After the system reaches a steady state, gravity is applied in six axis-aligned directions, each with a magnitude of 9.8 m/$\text{s}^2$ for 0.1 seconds.  A grasp pose is considered stable if the object remains in contact with the gripper at the final time step of the trajectory and if its center of mass moves less than a threshold proportional to $\epsilon_{v}\Delta t.$

\section{EXPERIMENTS}
\subsection{Multi-environment IPC Dataset Generation}
\label{sec:exp_parallel_env}

We first evaluate the performance of our parallel IPC simulator on a single NVIDIA H100 GPU (30 FP64 TFLOPS). Specifically, we run the dataset validation pipeline described in Section \ref{sec:grasp_validation} for rigid manipulands under varying numbers of parallel environments, ranging from a single environment to 512 environments. The speedup of $N$ parallel environments is measured as the runtime ratio compared to running $N$ environments sequentially.

The results in Fig. \ref{fig:scalability} demonstrate the strong performance of our approach: unimanual Leap Hand data generation achieves a 16× speedup with 512 environments, while unimanual UMI gripper data generation achieves a 40× speedup with 400 environments, enabling highly efficient large-scale data generation in parallel. To generate the full dataset, we deploy our automatic pipeline across 8 H100 GPUs, requiring approximately 600 GPU hours for all candidate poses, a process that would be prohibitively time-consuming without parallel environment support.

\subsection{Real2Sim Validation}
We select four objects of varying sizes and shapes from the single-hand manipuland set and fabricate them using a 3D printer with TPE-83A material. These objects are then grasped by a UMI gripper, and the resulting trajectories are recorded with a calibrated RealSense D435 camera. ArUco markers are attached to the manipulands and the grippers for 6D pose detection~\cite{marker_pose_detection}. The recorded gripper trajectories are subsequently replayed in our simulator, with the initial poses of the manipulands aligned to those in the real-world setup. As shown in Fig. \ref{fig:real2sim_exp1}, the bending behavior of the UMI gripper fingers and the 6D poses of the manipulands in the simulation closely match the real-world results, demonstrating high physical accuracy and fidelity.

\begin{figure}[t]
    \centering
    \includegraphics[width=0.8\linewidth]{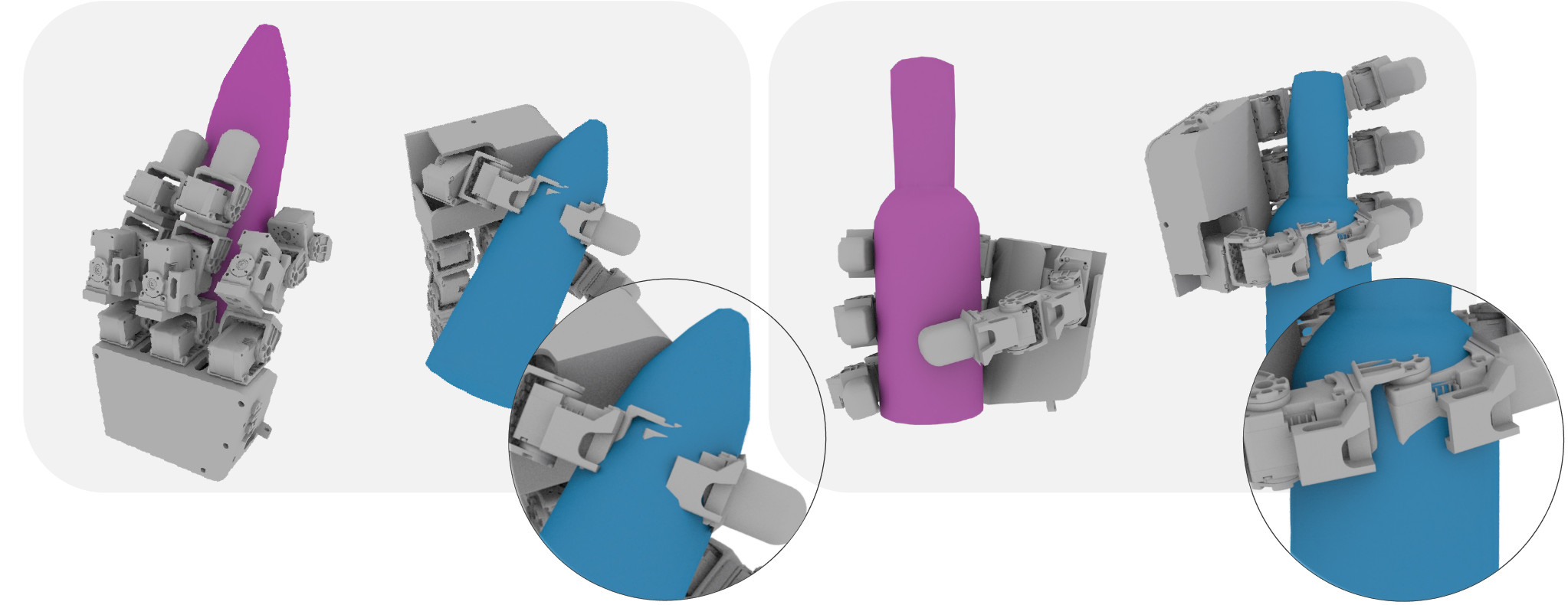}
    \caption{\textbf{IPC v.s. Isaac Gym Validation:} We filter the candidate poses generated by DexGraspNet \cite{wang2023dexgraspnet} using both our IPC simulator and the Isaac Gym \cite{makoviychuk2021isaac} simulator. Many poses deemed reasonable by Isaac Gym exhibit artificial penetrations, undermining validity. In contrast, our IPC simulator provides accurate, physically valid grasping behaviors with precise frictional contact modeling, effectively identifying and retaining only high-quality grasps.}
    \label{fig:isaacVSipc}
\end{figure}

\begin{figure}[b]
\centering
\includegraphics[width=0.6\linewidth]{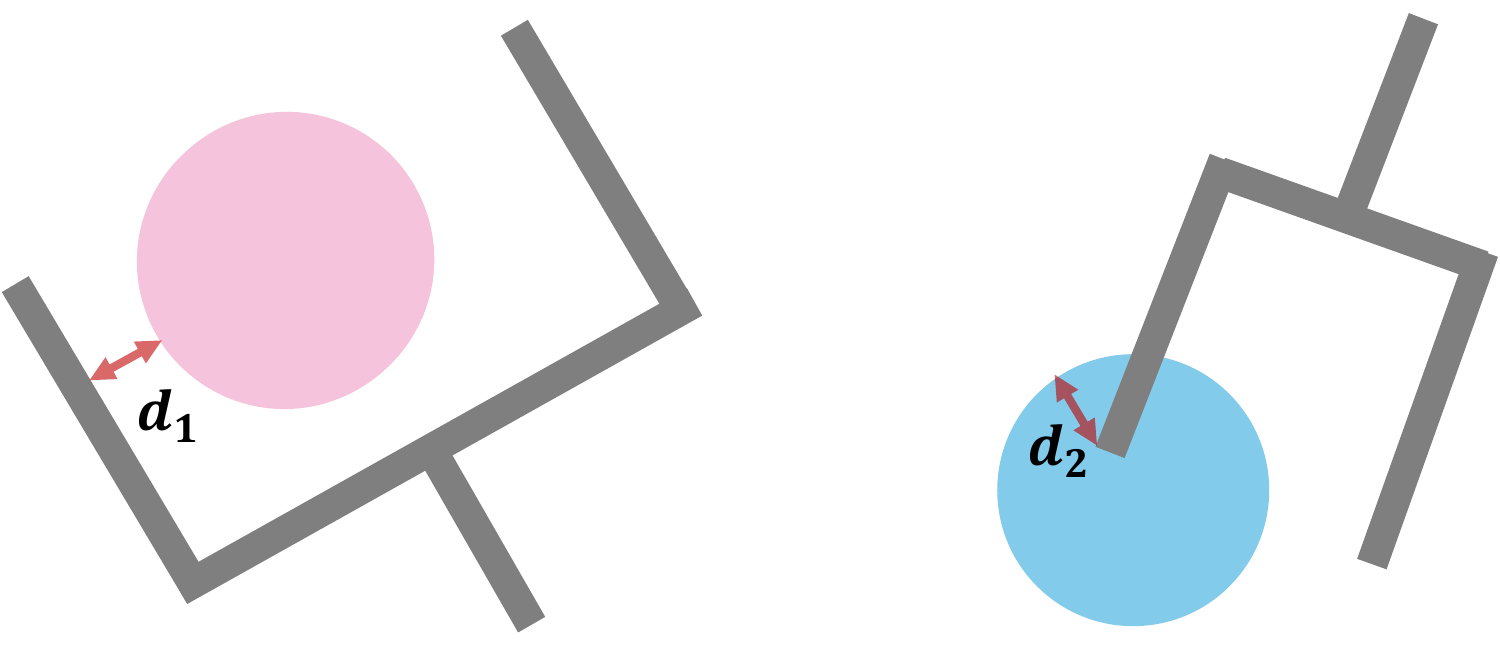}
\caption{\textbf{Illustration of Grasp Quality Metrics}: (\textbf{Left}) The gripper and manipuland are not in contact. In this case, $D_1=0$ and $D_2$ is the unsigned gripper-manipuland distance, $d_1$. (\textbf{Right}) Penetration occurs. Both $D_1$ and $D_2$ equal the penetration distance, $d_2$.}
\vspace{-2mm}
\label{fig:grasp_quality_metrics}
\end{figure}

\subsection{Dataset Quality Evaluation}
\paragraph{Quantitative Analysis}
At the core of our approach is an accurate and robust IPC-based simulator, designed to generate physically realistic grasp poses. To assess its effectiveness, we compare it with Isaac Gym \cite{isaac_gym}, a widely used simulator that is also adopted in DexGraspNet \cite{li2023gendexgrasp}. For a fair comparison, we use the same pipeline outlined in Fig.~\ref{fig:pipeline} but replace our IPC simulator with Isaac Gym for grasp pose validation. We then compare the quality of final grasp poses produced by both simulators, ensuring that the gripper securely holds the manipulands. 

To quantitatively analyze the grasp pose quality, we employ two metrics: (1) \emph{penetration distance} $D_1$, which measures the penetration depth between the object and the gripper, if any, and (2) \emph{absolute distance} $D_2$, defined as the unsigned distance between the object and the gripper to assess the grasp tightness. Fig.~\ref{fig:grasp_quality_metrics} illustrates the computation of these two metrics.
Both metrics are computed using a Signed Distance Field (SDF) $d^{\text{o}}:~\mathbb{R}^3 \rightarrow \mathbb{R}$ constructed for the manipuland, with point-SDF distances computed by sampling 50,000 points (forming a set $S^{\text{h}}$) on the LEAP Hand's surface:
\begin{align}
    D_1 &= \text{max}(0, \text{max}_{\vf{p}\in S^{\text{h}}}d^{\text{o}}(\vf{p}) ), \\
    D_2 &= |\text{max}_{\vf{p}\in S^{\text{h}}}d^{\text{o}}(\vf{p})|.
\end{align}

Using these two metrics, we evaluate the grasp quality produced by two simulators on a bottle subset (111 bottles) and the full 800-object dataset. As shown in Table \ref{tab:metrics}, our method achieves zero penetration cases, whereas Isaac Gym exhibits significant penetration artifacts (see Fig.~\ref{fig:isaacVSipc}), with an average penetration depth of 5 mm on the bottle subset and 8.7 mm across the full dataset. Additionally, the absolute distance in our method closely aligns with the IPC distance threshold parameter ($\hat{d}=10^{-3}$ m) used for surface compliance, indicating that the resulting grasps are tightly secured. In contrast, poses generated with Issac Gym exhibit significantly larger absolute distances, suggesting that the gripper may fail to hold the manipuland firmly. These results demonstrate that our IPC-based simulator not only eliminates penetration artifacts but also achieves stable grasps, making it a more reliable choice for a high-quality grasping dataset.

\begin{table}[t]
\centering
\setlength{\tabcolsep}{4pt}
\begin{tabular}{l|ccc|ccc}
\toprule
\multirow{2}{*}{\textbf{Simulator}} & \multicolumn{3}{c|}{\textbf{Bottle Subset}} & \multicolumn{3}{c}{\textbf{Full Dataset}} \\
& $D_1$ $\downarrow$ & $D_2$ $\downarrow$ & success $\uparrow$ & $D_1$ $\downarrow$ & $D_2$ $\downarrow$ & success $\uparrow$ \\
\midrule
\textbf{Ours} & \textbf{0} & \textbf{0.12} & \textbf{31.6\%} & \textbf{0} & \textbf{0.12} & \textbf{26.3\%} \\
Isaac Gym & 0.48 & 0.67 & 17.5\% & 0.87 & 1.19 & 7.9\% \\
\bottomrule
\end{tabular}
\caption{\textbf{Grasp Quality Benchmark}: Penetration distance $D_1$ and absolute distance $D_2$ are evaluated in centimeters (cm). The success rate measures whether the grasp poses generated by the trained UGG \cite{lu2024ugg} model can securely hold the manipuland within our IPC simulator.}
\vspace{-3mm}
\label{tab:metrics}
\end{table}

\begin{figure}[t]
    \centering
    \includegraphics[width=\linewidth]{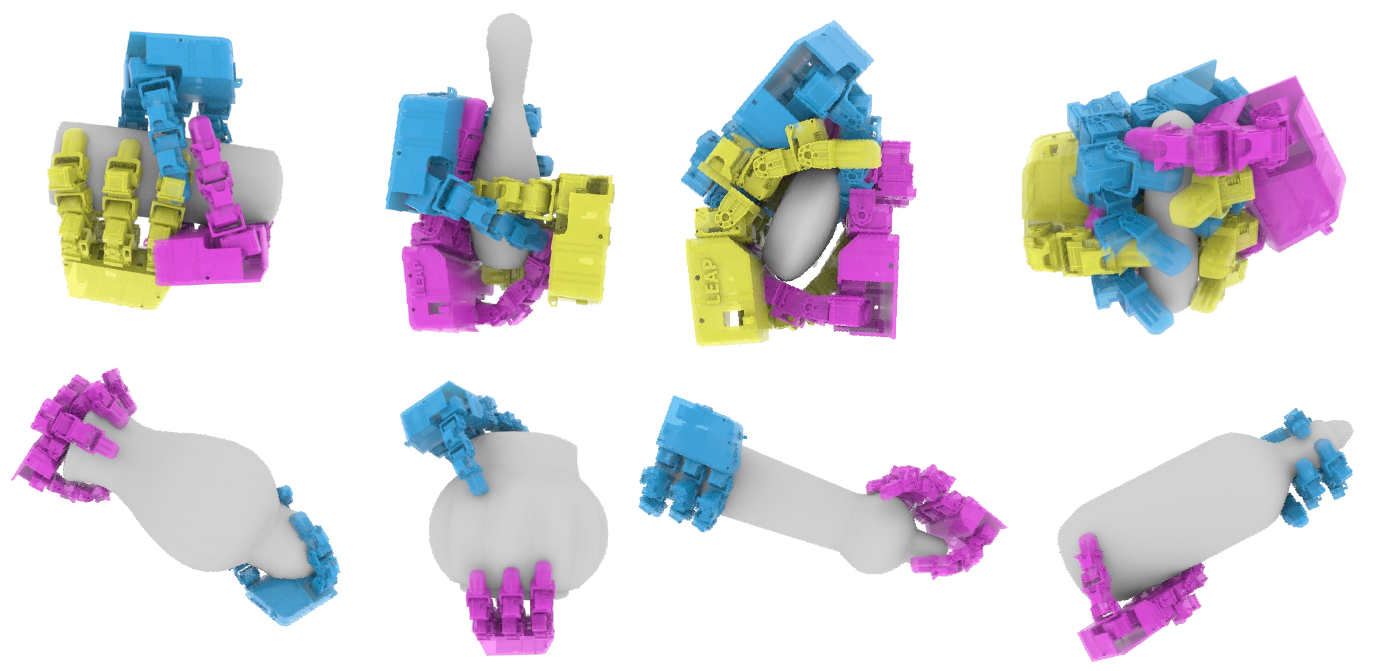}
    \caption{(\textbf{Top}) Neural grasp generation using UGG \cite{lu2024ugg} trained on our LEAP Hand unimanual dataset, demonstrating diverse grasp poses. (\textbf{Bottom}) The UGG network structure is modified to support bimanual grasp generation.}
    \label{fig:ugg_leap_bileap}
\end{figure}

\paragraph{Evaluation by Neural Grasp Generation}
We further evaluate our dataset on the downstream grasp generation task. Specifically, we train the diffusion-based dexterous grasp generation model Unified Generative Grasping (UGG) \cite{lu2024ugg} on two different datasets — one filtered using our IPC simulator and the other using Isaac Gym. We then validate the grasp poses with our IPC simulator to determine whether the generated poses can securely hold the manipulands without penetration. Table~\ref{tab:metrics} presents the grasp success rate of UGG models trained on these two datasets. The model trained on our dataset achieves a success rate of 31.6\% on the bottle subset and 26.3\% on the full dataset, significantly outperforming the model trained on the Isaac Gym-filtered dataset. In the top row of Fig.~\ref{fig:ugg_leap_bileap}, we visualize diverse grasping results generated by the UGG model trained on our dataset, demonstrating its ability to generate physically realistic and stable grasps. Furthermore, we extend the original UGG \cite{lu2024ugg} model to support bimanual grasp generation and train it on two bimanual subsets: a \emph{bottle} subset containing 48 bottle objects and a \emph{dispenser} subset containing 50 dispenser objects. The generated bimanual results are showcased at the bottom of Fig. \ref{fig:ugg_leap_bileap}, further illustrating the versatility of our dataset in supporting complex grasping scenarios.

\subsection{Stress Force Prediction}
Accurate stress prediction is essential for grasping and manipulating deformable objects to prevent damage and plastic deformation. Our dataset provides stress distributions for each grasp pose, enabling robust supervised learning for stress prediction. To demonstrate this, we train a neural network to predict the von Mises stress of a deformable object grasped by a UMI gripper, using its point cloud and the gripper mesh at the grasp pose as input.

Our model architecture follows \cite{deform_contact}, where we employ a Graph Neural Network (GNN) to extract features from both the object and the gripper. 
Additionally, a cross-attention module is incorporated to effectively capture hand-object interactions, enhancing stress prediction accuracy.

For training and evaluation, we utilize a subset of our GRIP dataset, consisting of eight bowls and eight mugs. Each category is split into a training set of six objects and a test set of two objects. Every object instance includes approximately 200 to 250 grasping poses. Due to the thin shell-like structures of these objects, their 3D-printed models will exhibit pronounced deformation during grasping, making them well-suited for analysis in real-world experiments.

To evaluate our stress prediction network, we employ two metrics: the Relative Mean Absolute Error (Relative MAE) $\mathcal{E}_{\text{Relative MAE}}=\frac{\sum_{i=1}^n|x_i-y_i|}{\max_{i\in [n]} |y_i|}$, and Kullback–Leibler (KL) divergence $\mathcal{E}_{\text{KL}}=\sum_{i=1}^{n} \hat{y}_i \log(\frac{\hat{y}_i}{\hat{x}_i})$, where $x_i$ and $y_i$ represent pointwise von Mises stress of the prediction and the simulation results, respectively. 
$\hat{x}_i$ and $\hat{y}_i$ are the corresponding empirical probability distributions computed as
$\hat{x}_i=\frac{x_i}{\sum_{i=1}^n x_i}$ and $\hat{y}_i=\frac{y_i}{\sum_{i=1}^n y_i}$. As shown in Table \ref{tab:exp2_metrics}, our model achieves an average Relative MAE of approximately $7\%\sim9\%$ and an average KL divergence $<0.2$ in the test set, demonstrating strong consistency between the predicted stress values and the simulation results.

\begin{figure}[t]
\centering
\includegraphics[width=1.0\linewidth]{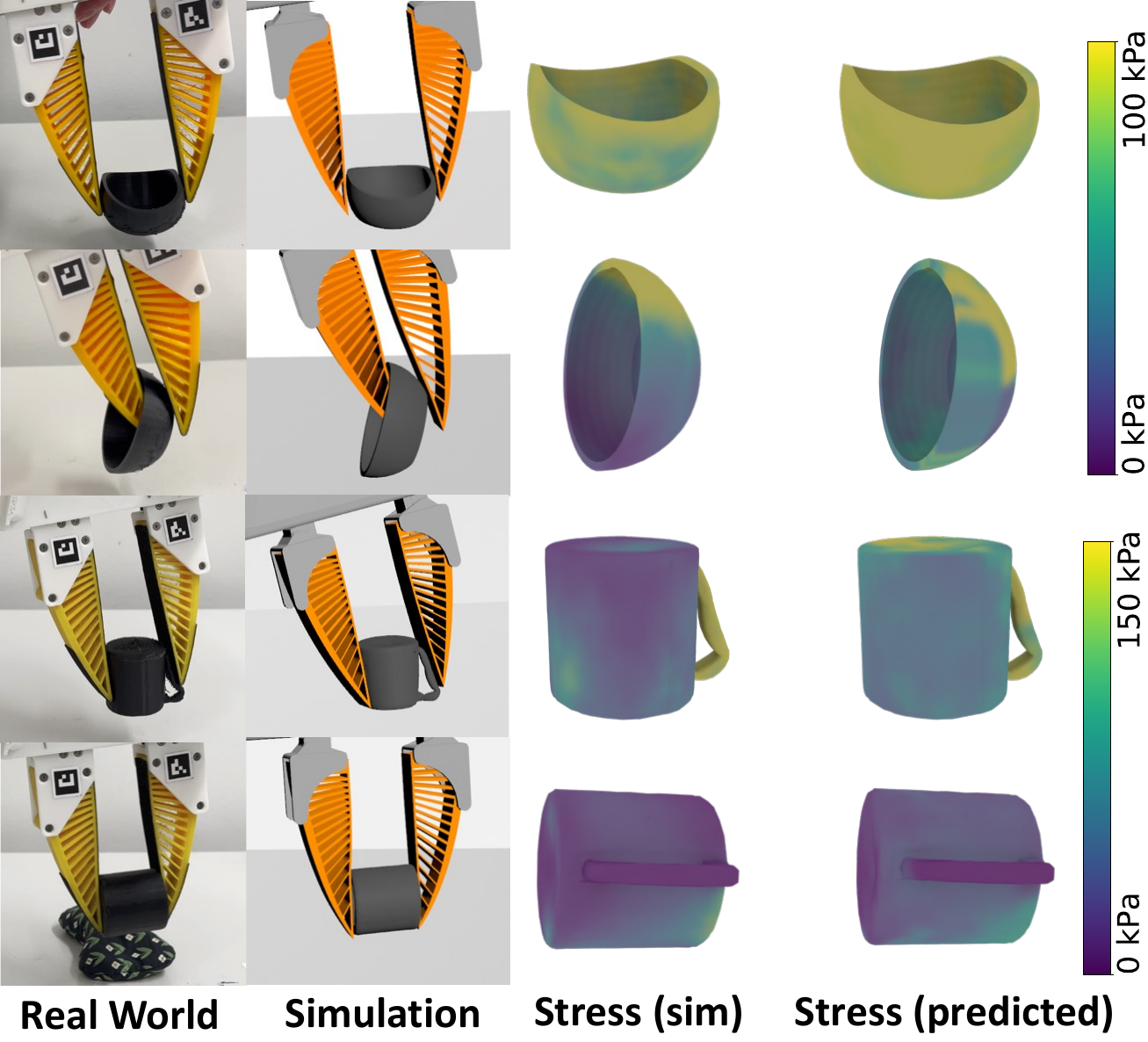}
\caption{\textbf{Stress Distribution Analysis on Real-world Grasps}: 
Each row shows a distinct grasp. The first and second columns display the grasps in the real world and simulation respectively. The third and fourth columns present the simulated and predicted normalized stress distributions.}
\label{fig:exp2_real_sim_pred_results}
\vspace{-3mm} 
\end{figure}

To further assess the accuracy and effectiveness of our stress prediction network, we apply the real-to-sim pipeline described in Section \ref{sec:exp1_real2sim} and conduct four grasping experiments using 3D-printed mug and bowl objects. Fig. \ref{fig:exp2_real_sim_pred_results} shows the normalized stress distribution predicted by our network alongside the simulation results, demonstrating strong agreement. 
For the grasp pose shown in the first row, the bowl undergoes significant deformation; rather, the grasp pose in the second row yields a moderate deformation. This aligns well with our network's predictions, where the mean stress in the first case is $2.17$ times that in the second. Similarly, for the two mug grasps, the predicted mean stress of the former is 1.6 times that of the latter, in accordance with visual deformation. Both the qualitative and quantitative evaluations confirm the high accuracy of our stress prediction network.

\section{CONCLUSION}
We introduce an IPC-based simulator optimized for large-scale, highly efficient, parallel multi-environment simulations, addressing key challenges in simulating soft grippers and deformable object grasping. Additionally, we develop a fully automated pipeline for grasp generation, simulation, and evaluation, supporting a wide range of grippers and manipuland types. Enabled by this pipeline, we construct the GRIP dataset, a large-scale, diverse grasp dataset featuring both soft and rigid grippers, to advance research in neural grasp generation and soft object manipulation.

Experimental results demonstrate the effectiveness of our
pipeline in synthesizing stable grasp poses and accurately
predicting manipuland stress distributions. Our current GRIP dataset features only UMI and LEAP Hands, but our gripper-agnostic pipeline can be easily extended to support diverse soft and rigid grippers. Through a real-to-simulation comparison, we further validate the strong agreement between our simulated results and real-world physics. However, our current grasp synthesis does not incorporate simulation feedback, as grasp generation is performed independently of the resulting dynamics. In future work, we aim to integrate differentiability into our pipeline to establish a closed-loop system, enabling dynamic feedback to refine grasp synthesis.

\begin{table}[t]
\centering
\begin{tabular}{lcc}
\toprule
Object Category  & Bowl & Mug  \\
\midrule
Relative MAE & 8.9\% &  7.2\% \\
KL divergence & 0.11 & 0.18 \\
\bottomrule
\end{tabular}
\caption{Relative Mean Absolute Error (MAE) and Kullback–Leibler (KL) divergence of our stress prediction network on the test set (a subset of the GRIP dataset) for the Bowl and Mug categories. }
\vspace{-3mm}
\label{tab:exp2_metrics}
\end{table}

\bibliographystyle{IEEEtran}
\bibliography{references}

\end{document}